\documentclass[letterpaper, 10 pt, conference]{ieeeconf}

\IEEEoverridecommandlockouts

\usepackage{amsmath}
\usepackage{amssymb}
\usepackage{soul}
\usepackage{booktabs}
\usepackage{graphicx}
\usepackage{algorithm}
\usepackage{algorithmic}

\usepackage{amsthm}
\usepackage[dvipsnames]{xcolor}
\usepackage{tikz}
\usepackage[normalem]{ulem}
\usepackage{subcaption}
\usepackage{multirow}
\usepackage{siunitx}
\usepackage{wasysym}  
\usepackage{colortbl}
\usepackage{pifont}
\definecolor{goodbg}{RGB}{230,245,234}

\newcommand{\ours}{\textsc{Ours}}

\usepackage{balance}
\usepackage{cite} 
\usepackage{hyperref}
\expandafter\def\expandafter\UrlBreaks\expandafter{\UrlBreaks\do\-}
\usepackage{float}

\definecolor{best}{RGB}{200, 230, 201}
\definecolor{secondbest}{RGB}{225, 245, 214}

\title{\LARGE \bf
Dense Temporal Motion Retargeting for Legged Robots
}

\author{Jaeryeong~Kim$^{1,*}$,
        Taerim~Yoon$^{1,*}$,
        Jin~Cheng$^{1}$,
        Sungjoon~Choi$^{2}$, and
        Stelian~Coros$^{1}$
\thanks{$^{*}$Equal contribution.
Project page: {\urlstyle{same}\url{https://jaeryeongnicolekim.com/Dense-Temporal-Motion-Retargeting-For-Legged-Robots/}}.
}
\thanks{$^{1}$Jaeryeong~Kim, Taerim~Yoon, Jin~Cheng, and Stelian~Coros are with the Department of Computer Science, ETH Zurich, Wasserwerkstrasse 12, 8092 Zurich, Switzerland (email: jaerkim@ethz.ch, tayoon@ethz.ch, jicheng@ethz.ch, scoros@ethz.ch).}
\thanks{$^{2}$Sungjoon~Choi is with the Department of Artificial Intelligence, Korea University, 145 Anam-ro, Seongbuk-gu, Seoul, Korea (email: sungjoon-choi@korea.ac.kr).}
}

\begin{document}

\maketitle
\thispagestyle{empty}
\pagenumbering{Arabic}

\begin{abstract}
Legged robots can learn expressive whole-body skills from the motions of humans and animals.
Due to the morphology gap between the source and the robot, however, the motion must be tailored to the dynamic properties of the robot.
In particular, dynamic motions such as a jump require careful adjustment, since their timing and control are interdependent.
We propose \emph{dense temporal motion retargeting} (DTMR), which jointly optimizes timing and control within a single optimization, where dense means that the timing is adjusted for every control step.
This dense formulation enables DTMR to deform only the parts of the motion that need a change in timing.
The problem is solved with sampling-based model predictive control (MPC) in parallel on a GPU.
We evaluate DTMR against baselines on two hours of human motion with four humanoid robots, where the results show that DTMR outperforms baseline methods, particularly on dynamic motions.
We also show that allowing more temporal deformation yields more precise retargeting.
We further compare DTMR with a baseline that optimizes the temporal dimension, where the result shows that DTMR retargets more precisely under the same deformation budget while being ${\sim}19\times$ faster.
Lastly, policies trained on our references transfer to a real humanoid robot.
\end{abstract}

\section{Introduction}\label{sec:intro}

Legged robots have bodies like those of biological creatures, which lets them learn from the motions of humans and animals~\cite{Serifi2024VMPVM, ze2025twist, luo2025sonic}.
Such references are abundant and semantically rich, but the morphological and dimensional gap between the source and the robot prevents their direct use.
\emph{Motion retargeting} closes this gap by adapting the source motion to the target robot~\cite{araujo2025gmr, lee2025phuma, yang2025omniretarget}.
However, the robot can imitate the retargeted motion only if it is dynamically feasible~\cite{luo2023phc}.
For instance, a jumping motion requires careful tailoring of both timing and control to the kino-dynamic properties of the robot.
Our goal is to retarget such dynamic motions with their timing adjusted so that legged robots can execute them.

The main challenge is that control and timing are interdependent.
On one hand, the motion should be adapted to what the robot can do.
For instance, a robot with weaker actuators cannot move as fast and must slow the motion down.
On the other hand, the control must change with the adapted motion.

\begin{figure}[t]
  \centering
  \includegraphics[width=\linewidth]{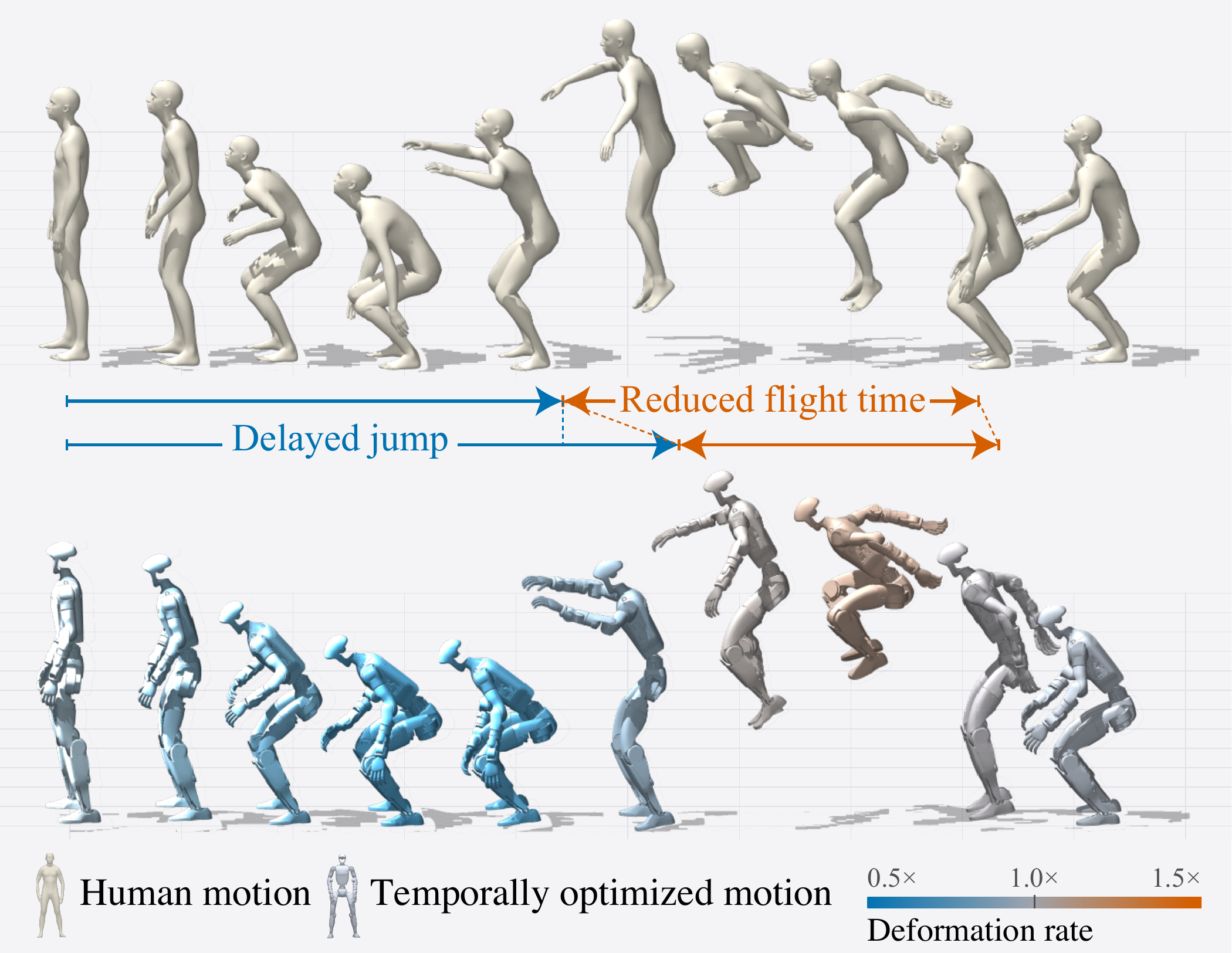}
  \caption{The timing of a human jump is adjusted according to the dynamic
    properties of the Unitree G1. The take-off is delayed (blue) and the
    flight is sped up (orange).}
  \label{fig:teaser}
\end{figure}

To overcome this, \emph{temporal motion retargeting} has been introduced, which adapts the timing to the dynamics of the robot~\cite{yoon2025spatio}.
However, the existing approach has two limitations.
First, it adjusts the timing per motion segment.
This deforms frames that need no change, since the whole segment is scaled.
Second, the timing is found by a slow nested optimization that solves the control problem repeatedly.

To this end, we propose \emph{dense temporal motion retargeting} (DTMR) that optimizes dense timing and control within a single optimization, where \emph{dense} refers to adjusting the timing at every frame rather than per motion segment.
Only the parts of the motion that need a change are deformed, such as the take-off and the flight in Fig.~\ref{fig:teaser}.
We show that the resulting problem can be solved efficiently with sampling-based model predictive control (MPC)~\cite{williams2018information} by leveraging GPU parallelization.

We evaluate our method on a two-hour human motion dataset~\cite{bones_seed_2026}.
We demonstrate that our method retargets motions more successfully than baselines when evaluated through downstream imitation learning.
In particular, temporal optimization is especially effective on dynamic motions.
We also demonstrate that allowing more temporal deformation leads to more precise retargeting, which gives an effective knob between timing preservation and precision.
Furthermore, DTMR retargets more precisely than baseline temporal optimization~\cite{yoon2025spatio} under a similar deformation budget.
Finally, policies trained on our references transfer to a real humanoid robot.

Our contributions are summarized as follows:
\begin{itemize}
  \item We propose DTMR, which optimizes the timing and control of every frame together and deforms only the parts of the motion that need a change.
  \item We show that DTMR can allow more temporal deformation for more precise retargeting, and that under the same deformation budget it retargets more precisely than the baseline and ${\sim}19\times$ faster.
  \item We construct a two-hour dataset of dynamically feasible motions, on which policies learn dynamic motions better and transfer to a real humanoid robot.
\end{itemize}

\section{Related Work}\label{sec:related_work}

To imitate the motion of a biological creature, a legged robot must overcome the morphology gap between the source and the target.
Since the retargeted motion is then imitated by a downstream imitation learning method~\cite{peng2018deepmimic, ze2025twist, chen2025gmt, luo2025sonic}, the quality of the retargeting is crucial.

One line of work addresses this gap at the kinematic level.
Yamane et al.~\cite{yamane2010animating} learned a pose-to-pose mapping from paired motion data, which requires a costly data collection for every new pair of embodiments.
To remove this requirement, Choi and Ko~\cite{choi2000online} transferred keypoint trajectories through inverse kinematics, and Aberman et al.~\cite{aberman2020skeleton} learned the correspondence without paired data through a shared skeleton representation.
Recent methods further add constraints for physical plausibility, such as the interaction mesh of Yang et al.~\cite{yang2025omniretarget} for contacts with objects and the terrain, and the joint limits and ground contact of Lee et al.~\cite{lee2025phuma}.
These methods produce spatially plausible poses, but they do not account for the dynamics of the robot.

A second line of work incorporates the dynamics of the target into the retargeting.
Tak and Ko~\cite{tak2005physically} filtered the motion with balance constraints, Rouxel et al.~\cite{rouxel2022multicontact} solved a whole-body optimization with contact forces, and Grandia et al.~\cite{grandia2023doc} differentiated through optimal control to retarget motions onto legged robots.
These methods make the motion dynamically feasible, but they keep the timing of the source, which the robot may not be able to follow.

Temporal retargeting has been studied to address this limitation.
Yoon et al.~\cite{yoon2025spatio} scaled the time of a few coarse motion segments to obtain dynamically feasible motions for quadruped robots.
The time scaling is found by a nested optimization where the outer loop searches the temporal parameters with Bayesian optimization~\cite{snoek2012practical} and the inner loop finds the control with iLQG~\cite{tassa2012synthesis}.
However, the nested optimization is slow, since every outer iteration requires a full inner solve, and the coarse scaling deforms frames that need no change.
In contrast, we optimize the timing of every frame together with the control in a single MPPI problem.
As a result, the feasible portions of the motion are left intact and only the infeasible portions are locally reparameterized in time.

\section{Preliminaries}\label{sec:prelim}

MPPI~\cite{williams2018information} is a sampling-based model predictive controller.
At each control step $k$ it updates a plan $\mathbf{U}=\mathbf{u}_{k:k+H}=(\mathbf{u}_k,\dots,\mathbf{u}_{k+H})$ over a horizon $H$ from the current state, applies $\mathbf{u}_k$, and shifts the plan by one step to warm-start step $k+1$.
The update treats the applied input as a noisy version of the plan, $\mathbf{v}_{k+h}=\mathbf{u}_{k+h}+\boldsymbol{\epsilon}_h$ with $\boldsymbol{\epsilon}_h\sim\mathcal{N}(\mathbf{0},\boldsymbol{\Sigma})$, $h=0,\dots,H$.
Given a trajectory cost $J(\mathbf{V})=\sum_h c(\mathbf{x}_{k+h},\mathbf{v}_{k+h})$ evaluated by rolling out $f$, sampling $N_W$ noise sequences and weighting each rollout by its cost gives the update
\begin{equation}
  \mathbf{U}\leftarrow\mathbf{U}+\sum_{n=1}^{N_W}w_n\,[W]_n,\qquad
  w_n=\frac{e^{-J_n/\lambda}}{\sum_{m}e^{-J_m/\lambda}},
  \label{eq:mppi}
\end{equation}
with temperature $\lambda$, $[W]_{1:N_W}$ the $N_W$ sampled noise sequences, each $[W]_n=(\boldsymbol{\epsilon}_0,\dots,\boldsymbol{\epsilon}_H)$, and $J_n=J(\mathbf{U}+[W]_n)$ its rollout cost.
Eq.~\eqref{eq:mppi} is applied once per control step.
Since the $N_W$ rollouts are independent, the update is efficiently parallelized on a GPU.
In this work, we use MPPI to optimize the motor control and the progress through the source motion simultaneously as a single control signal.

\section{Problem Formulation}\label{sec:problem}
Consider a legged robot with state $\mathbf{x}_k=(\mathbf{q}_k,\dot{\mathbf{q}}_k)$, where $\mathbf{q}_k$ collects the root pose and the joint angles, and motor control $\mathbf{u}_k$.
Its dynamics are described as $\mathbf{x}_{k+1}=f(\mathbf{x}_k,\mathbf{u}_k)$ with a discrete control period $\Delta t$.
The robot is given a source motion of duration $T$ as a trajectory $\bar{\mathbf{p}}$ of body keypoints.

We define a phase variable $\phi\in[0,1]$ that maps to the source motion from its beginning to its end.
The phase denotes the point currently being referenced in the source motion, and its increment determines how quickly the robot advances through the motion.
For instance, a smaller increment slows down a portion that the actuators of the robot cannot follow at the source speed.

Motion retargeting seeks the motor controls under which the robot reproduces the source motion.
Temporal motion retargeting additionally lets the timing change.
The phase $\phi_k$ at control step $k$ is itself a decision variable, so that the robot may imitate the source faster or slower than the source itself.
This can be described as
\begin{equation}
  \begin{aligned}
    \min_{\mathbf{u},\boldsymbol{\phi}}\;\; & \frac{1}{2T_{\phi}}
    \sum_{k=0}^{T_{\phi}}
    \big\|\mathrm{FK}(\mathbf{x}_k)-\mathrm{LI}(\phi_k;\bar{\mathbf{p}})\big\|_Q^2 \\[2pt]
    \text{s.t.}\;\;                                    & \mathbf{x}_{k+1}=f(\mathbf{x}_k,\mathbf{u}_k),\\
                                                       & \phi_0=0,\quad \phi_k\le\phi_{k+1},\quad \phi_{T_\phi}=1,
  \end{aligned}
  \label{eq:tmr}
\end{equation}
where $\mathrm{LI}(\phi_k;\bar{\mathbf{p}})$ denotes the keypoints of the source motion at phase $\phi_k$ obtained by linear interpolation between the neighboring frames, $\mathrm{FK}(\mathbf{x}_k)$ the corresponding keypoints of the robot from forward kinematics, $Q$ a weight matrix, and $T_{\phi}$ the number of control steps until $\phi$ reaches~1.
The phase trajectory $\boldsymbol{\phi}=\{\phi_k\}$ decides which part of the source the robot imitates at each control step, and the motor controls $\mathbf{u}$ must realize it under the dynamics of the robot.
Eq.~\eqref{eq:tmr} therefore searches the timing and the motor control together.

\section{Method}\label{sec:method}

Our goal is to search the timing and the motor control of Eq.~\eqref{eq:tmr} simultaneously.
We define the phase-augmented state, control, and dynamics
\begin{equation*}
  \tilde{\mathbf{x}}_k=(\mathbf{x}_k,\phi_k),\quad
  \tilde{\mathbf{u}}_k=(\mathbf{u}_k,d\phi_k),\quad
  \tilde{\mathbf{x}}_{k+1}=\tilde f(\tilde{\mathbf{x}}_k,\tilde{\mathbf{u}}_k),
\end{equation*}
where $\tilde f$ advances the robot and the phase as $\mathbf{x}_{k+1}=f(\mathbf{x}_k,\mathbf{u}_k)$ and $\phi_{k+1}=\phi_k+d\phi_k$.
The phase increment $d\phi_k$ is a control input.
With $d\phi^{\mathrm{src}}=\Delta t/T$ the increment at the source speed, the deformation rate $r_k=\log_2(d\phi_k/d\phi^{\mathrm{src}})$ is negative where the robot slows down and positive where it speeds up, so that $-1$ means twice as slow and $1$ twice as fast.
In this way, the phase trajectory $\boldsymbol{\phi}$ of Eq.~\eqref{eq:tmr} is no longer a separate variable but is generated by the control sequence through $d\phi_k$.
Timing and motor control are thus optimized in one control problem.
DTMR is then the single optimization
\begin{equation}
  \begin{aligned}
    \min_{\tilde{\mathbf{u}}}\;\; & \frac{1}{T_\phi}\sum_{k=0}^{T_\phi}
    c(\tilde{\mathbf{x}}_k,\tilde{\mathbf{u}}_k)                                                                  \\[2pt]
    \text{s.t.}\;\;               & \tilde{\mathbf{x}}_{k+1}=\tilde f(\tilde{\mathbf{x}}_k,\tilde{\mathbf{u}}_k),
  \end{aligned}
  \label{eq:dense}
\end{equation}
where $T_\phi=\min\{k:\phi_k=1\}$ is the number of control steps until the rollout reaches the end of the motion, determined by the rollout rather than chosen in advance.

The running cost of Eq.~\eqref{eq:dense} is
\begin{equation}
  c(\tilde{\mathbf{x}}_k,\tilde{\mathbf{u}}_k)=
  c_{\mathrm{track}}+w_{d\phi}\,c_{\phi}+c_{\mathrm{reg}},
  \label{eq:cost}
\end{equation}
evaluated after the physics step and the phase increment, so the tracking term compares the robot with the reference frame that $d\phi_k$ selected.
Table~\ref{tab:cost} lists the terms, and the overall procedure is summarized in Alg.~\ref{alg:dense_tmr}.

\begin{algorithm}[!t]
  \caption{Dense Temporal Motion Retargeting}
  \label{alg:dense_tmr}
  \small
  \begin{algorithmic}[1]
    \STATE Initialize $\tilde{\mathbf{u}}_{0:H}\gets(\bar{\mathbf{q}}_{0:H},\,d\phi^{\mathrm{src}})$
    \STATE $k\gets0$, $\phi_0\gets0$
    \WHILE{$\phi_k<1$}
    \FOR{$i=1$ \TO $N$}
    \STATE Get noise kernel $\tilde\Sigma^{i}_{k:k+H}=\mathrm{diag}\big(\Sigma^{i}_{k:k+H},\,\sigma^2_{\phi,i}\big)$, $\Sigma^{i}$ from the annealing schedule of DIAL-MPC~\cite{xue2024dialmpc}
    \STATE Sample $[\tilde W]_{1:N_W}\sim\mathcal{N}(\mathbf{0},\tilde\Sigma^{i}_{k:k+H})$
    \STATE Rollout $\tilde f$ from $(\mathbf{x}_k,\phi_k)$ under $\tilde{\mathbf{u}}_{k:k+H}+[\tilde W]_{1:N_W}$, advancing $\phi$ before each physics step, and evaluate $J(\cdot)$ with Eq.~\eqref{eq:cost}
    \STATE Keep the elite fraction $\kappa$ of the rollouts with the lowest cost
    \STATE Update $\tilde{\mathbf{u}}^{(i)}_{k:k+H}$ with Eq.~\eqref{eq:mppi} over the elite rollouts
    \ENDFOR
    \STATE $(\mathbf{u}_k,d\phi_k)\gets\tilde{\mathbf{u}}_k$
    \STATE $\phi_{k+1}\gets\min(\phi_k+d\phi_k,1)$
    \STATE $\mathbf{x}_{k+1}\gets f(\mathbf{x}_k,\mathbf{u}_k)$
    \STATE Shift the plan $\tilde{\mathbf{u}}_{k+1:k+H+1}\gets\mathtt{shift}(\tilde{\mathbf{u}}_{k:k+H})$ and fill the last entry with $(\bar{\mathbf{q}}_{k+H+1},\,d\phi^{\mathrm{src}})$
    \ENDWHILE
    \RETURN refined reference $\{\mathbf{x}_k\}$ and phase trajectory $\{\phi_k\}$
  \end{algorithmic}
\end{algorithm}

\begin{table}[t]
  \centering
  \caption{The cost terms are listed with their weights.}
  \label{tab:cost}
  \small
  \begin{tabular}{@{}lll@{}}
    \toprule
    Term                                    & Expression                                                             & Weight \\
    \midrule
    root position ($c_{\mathrm{track}}$)    & $\rho(\mathbf{p}^{\mathrm{root}}-\bar{\mathbf{p}}^{\mathrm{root}})$    & 28     \\
    link position ($c_{\mathrm{track}}$)    & $\sum_\ell s_\ell\,\rho(\Delta\mathbf{p}_\ell)$                        & 14     \\
    link orientation ($c_{\mathrm{track}}$) & $\sum_\ell s_\ell\,\rho(\Delta\mathbf{R}_\ell)$                        & 5      \\
    joint velocity ($c_{\mathrm{track}}$)   & $\tfrac12\|\dot{\mathbf{q}}_k-2^{r_k}\dot{\bar{\mathbf{q}}}(\phi_k)\|^2$ & 0.01   \\
    phase ($c_\phi$)                        & $\|r_k\|^2_{Q_\phi}$             & 5      \\
    torque energy ($c_{\mathrm{reg}}$)      & $\sum_i\big[\cosh(\tau_{k,i}/\tau_i^{\mathrm{rng}})-1\big]$           & 0.05   \\
    action rate ($c_{\mathrm{reg}}$)        & $\|\mathbf{u}_k-\mathbf{u}_{k-1}\|^2$                                  & 1.0    \\
    \bottomrule
  \end{tabular}
\end{table}

The tracking term $c_{\mathrm{track}}$ penalizes the error between the robot and the reference keypoints at phase $\phi_k$, namely the root position in the world frame, the root-relative position error $\Delta\mathbf{p}_\ell$ of each link $\ell$, and its orientation error $\Delta\mathbf{R}_\ell=\mathbf{R}_\ell-\bar{\mathbf{R}}_\ell$.
Errors are measured with the robust norm $\rho(\mathbf{e})=\sum_i[(|e_i|^q+\delta^q)^{1/q}-\delta]$, which ignores residuals within a margin $\delta$ and grows linearly beyond, with a larger weight $s_\ell=16$ on the torso and head and $1$ otherwise; $\delta$ and $q$ are listed in Table~\ref{tab:hparams}.
Since the reference is played back at $2^{r_k}$ times the source speed, its joint velocities are multiplied by $2^{r_k}$ before being used as the tracking target.
The regularization term $c_{\mathrm{reg}}$ penalizes torque energy, normalized by the torque range $\tau_i^{\mathrm{rng}}$ of each actuator, and action rate.
No contact schedule or feasibility constraint is needed, because every candidate is a simulated rollout.

The phase term $c_\phi$ in Table~\ref{tab:cost} penalizes deviation from the source speed, which is zero at the source speed and quadratic in the deformation rate, so one weight transfers across clips of differing tempo.
The weight $w_{d\phi}$ controls the trade-off between the amount of temporal deformation and the retargeting precision, as demonstrated in Sec.~\ref{subsec:eval_tradeoff}.
Note that we set $Q_\phi$ to zero for $d\phi_k>d\phi^{\mathrm{src}}$ and penalize slow-down only, since empirically slowing down occurs far more readily than speeding up.

\begin{table}[!t]
  \centering
  \caption{List of the hyperparameters used for DTMR.}
  \label{tab:hparams}
  \small
  \begin{tabular}{@{}ll@{}}
    \toprule
    Setting                                 & Value                       \\
    \midrule
    samples ($N_W$)                         & 4096                        \\
    horizon ($H$)                           & 40 steps                    \\
    control nodes ($H_{\mathrm{node}}$)     & 8                           \\
    annealing iterations ($N$, cold / warm) & 5 / 2                       \\
    noise ($\sigma_0$), decay ($\beta$)     & 0.15, 0.6                   \\
    horizon envelope ($\beta_h$)            & 0.95                        \\
    temperature ($\lambda$)                 & 0.3                         \\
    elite fraction ($\kappa$)               & 0.1                         \\
    phase noise ($\sigma_\phi$)             & 0.3                         \\
    robust norm ($\delta$, $q$), root       & 0.1\,m, 2                   \\
    robust norm ($\delta$, $q$), links      & 0.2\,m, 4                   \\
    control / physics step                  & 20 / 5\,ms                  \\
    PD gains ($k_p$, $k_d$), G1             &                             \\
    \quad hip, knee, ankle                  & (150, 2), (200, 4), (40, 2) \\
    \quad waist                             & (250, 5)                    \\
    \quad shoulder, elbow, wrist            & (100, 2), (40, 2), (20, 1)  \\
    PD gains ($k_p$, $k_d$), Go1            & (30, 1)                     \\
    \bottomrule
  \end{tabular}
\end{table}

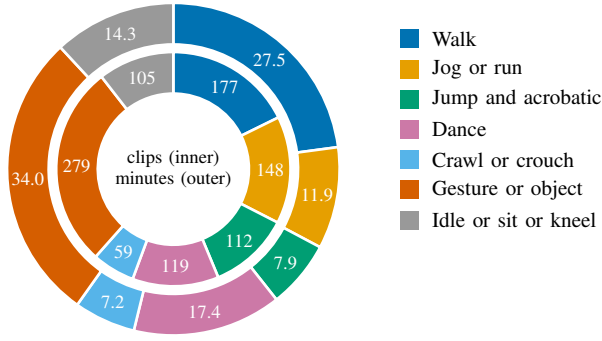
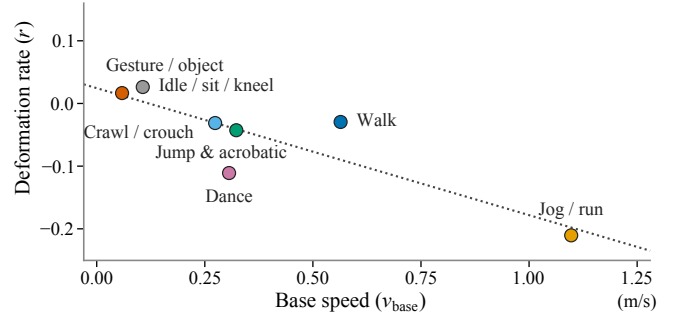
\begin{figure*}[t]
  \centering
  \begin{subfigure}[b]{0.49\textwidth}
    \centering
    \definecolor{DC1}{HTML}{0072B2}\definecolor{DC2}{HTML}{E69F00}\definecolor{DC3}{HTML}{009E73}
\definecolor{DC4}{HTML}{CC79A7}\definecolor{DC5}{HTML}{56B4E9}\definecolor{DC6}{HTML}{D55E00}\definecolor{DC7}{HTML}{999999}
\def\donutAc{90}\def\donutAm{90}
\begin{tikzpicture}[font=\footnotesize]
  \foreach \name/\c/\m/\col [count=\i] in {Walk/177/27.5/DC1, Jog or run/148/11.9/DC2, Jump and acrobatic/112/7.9/DC3, Dance/119/17.4/DC4, Crawl or crouch/59/7.2/DC5, Gesture or object/279/34.0/DC6, Idle or sit or kneel/105/14.3/DC7} {
    \pgfmathsetmacro{\endc}{\donutAc-\c/999*360}
    \fill[\col, draw=white, line width=1pt] (\donutAc:1.0) arc (\donutAc:\endc:1.0) -- (\endc:1.55) arc (\endc:\donutAc:1.55) -- cycle;
    \pgfmathsetmacro{\midc}{(\donutAc+\endc)/2}
    \node[white, font=\scriptsize] at (\midc:1.28) {\c};
    \pgfmathsetmacro{\endm}{\donutAm-\m/120.2*360}
    \fill[\col, draw=white, line width=1pt] (\donutAm:1.65) arc (\donutAm:\endm:1.65) -- (\endm:2.2) arc (\endm:\donutAm:2.2) -- cycle;
    \pgfmathsetmacro{\midm}{(\donutAm+\endm)/2}
    \node[white, font=\scriptsize] at (\midm:1.93) {\m};
    \global\edef\donutAc{\endc}\global\edef\donutAm{\endm}
    \fill[\col] (3.0,{2.0-0.4*\i}) rectangle ++(0.25,0.25);
    \node[anchor=west] at (3.3,{2.125-0.4*\i}) {\name};
  }
  \node[font=\scriptsize, align=center] at (0,0) {clips (inner)\\minutes (outer)};
\end{tikzpicture}
    \caption{Motion classes of the dataset.}
    \label{fig:seed2h_donut}
  \end{subfigure}
  \hfill
  \begin{subfigure}[b]{0.49\textwidth}
    \centering
    \includegraphics[width=\linewidth]{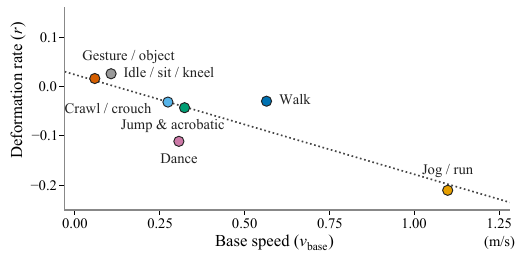}
    \caption{Re-timing per motion class on the G1.}
    \label{fig:seed2h_retiming}
  \end{subfigure}
  \caption{(a) The dataset retargeted from two hours of human motion, by the number of clips (inner ring) and the total duration in minutes (outer ring) per motion class.
    (b) 
    The mean deformation rate of each motion class on the G1 against the mean base speed of its reference clips. The more dynamic a class is, the more it is slowed down to fit the dynamics of the robot, while slow classes are sped up slightly, as the dotted trend line shows.}
  \label{fig:seed2h_classes}
\end{figure*}

\begin{table*}[t]
  \centering
  \caption{Dynamic feasibility of the retargeted motions is compared across baseline methods. The retargeting error is reported in mm at the 25, 50, and 75\,\% quartiles over clips. Lower is better, and the best value per column is in bold.}
  \label{tab:twist_training}
  \small
  \begin{tabular*}{\textwidth}{@{\extracolsep{\fill}}l ccc ccc ccc ccc@{}}
    \toprule
    & \multicolumn{3}{c}{G1} & \multicolumn{3}{c}{R1} & \multicolumn{3}{c}{H1-2} & \multicolumn{3}{c}{T1} \\
    \cmidrule(lr){2-4}\cmidrule(lr){5-7}\cmidrule(lr){8-10}\cmidrule(lr){11-13}
    Retargeting error (mm) $\downarrow$ & $Q_{25}$ & $Q_{50}$ & $Q_{75}$ & $Q_{25}$ & $Q_{50}$ & $Q_{75}$ & $Q_{25}$ & $Q_{50}$ & $Q_{75}$ & $Q_{25}$ & $Q_{50}$ & $Q_{75}$ \\
    \midrule
    GMR~\cite{araujo2025gmr} & 52.7 & 113.4 & 206.5 & 56.7 & 107.5 & 215.1 & 73.4 & 148.4 & 509.7 & 52.0 & 116.2 & 375.2 \\
    PHUMA~\cite{lee2025phuma} & 70.5 & 137.3 & 228.0 & 46.6 & 155.7 & 401.7 & 74.2 & 279.6 & 871.6 & 64.1 & 379.4 & 1627.7 \\
    OmniRetarget~\cite{yang2025omniretarget} & 59.3 & 98.4 & 153.9 & 55.7 & 94.0 & 150.7 & 84.4 & 257.4 & 748.8 & 105.0 & 339.6 & 1306.2 \\
    SoMa-RT~\cite{saito2026soma} & 54.8 & 103.7 & 157.6 & \textbf{45.0} & 83.9 & 140.9 & \textbf{68.2} & 151.8 & 492.6 & 50.8 & 132.0 & 403.7 \\
    SoMa-RT\,$+$\,DTMR (\ours) & \textbf{48.5} & \textbf{81.4} & \textbf{140.3} & 45.3 & \textbf{77.6} & \textbf{134.5} & 71.4 & \textbf{136.3} & \textbf{377.2} & \textbf{50.3} & \textbf{112.8} & \textbf{320.2} \\
    \bottomrule
  \end{tabular*}
\end{table*}

\begin{table*}[t]
  \centering
  \caption{Semantic fidelity of the retargeted motions is compared across baseline methods. To be insensitive to temporal deformation, the DTW retargeting error is reported in mm at the 25, 50, and 75\,\% quartiles over clips. Lower is better, and the best value per column is in bold.}
  \label{tab:twist_dtw}
  \small
  \begin{tabular*}{\textwidth}{@{\extracolsep{\fill}}l ccc ccc ccc ccc@{}}
    \toprule
    & \multicolumn{3}{c}{G1} & \multicolumn{3}{c}{R1} & \multicolumn{3}{c}{H1-2} & \multicolumn{3}{c}{T1} \\
    \cmidrule(lr){2-4}\cmidrule(lr){5-7}\cmidrule(lr){8-10}\cmidrule(lr){11-13}
    DTW retargeting error (mm) $\downarrow$ & $Q_{25}$ & $Q_{50}$ & $Q_{75}$ & $Q_{25}$ & $Q_{50}$ & $Q_{75}$ & $Q_{25}$ & $Q_{50}$ & $Q_{75}$ & $Q_{25}$ & $Q_{50}$ & $Q_{75}$ \\
    \midrule
    GMR~\cite{araujo2025gmr} & 51.0 & 102.7 & 169.5 & 54.4 & 96.7 & 179.8 & 70.4 & 135.7 & 442.0 & 51.3 & 107.0 & 354.7 \\
    PHUMA~\cite{lee2025phuma} & 66.6 & 119.0 & 184.0 & 45.7 & 141.8 & 318.7 & 73.6 & 258.5 & 686.3 & 63.6 & 373.3 & 1553.1 \\
    OmniRetarget~\cite{yang2025omniretarget} & 55.6 & 90.1 & 139.1 & 53.6 & 86.8 & 131.7 & 83.0 & 230.6 & 594.0 & 104.4 & 336.7 & 1299.4 \\
    SoMa-RT~\cite{saito2026soma} & 52.2 & 93.2 & 141.5 & 43.3 & 75.9 & 123.6 & \textbf{65.8} & 141.2 & 392.0 & 50.0 & 117.7 & 359.5 \\
    SoMa-RT\,$+$\,DTMR (\ours) & \textbf{45.8} & \textbf{74.1} & \textbf{128.6} & \textbf{43.1} & \textbf{70.4} & \textbf{121.1} & 69.3 & \textbf{127.4} & \textbf{333.1} & \textbf{49.3} & \textbf{101.2} & \textbf{289.1} \\
    \bottomrule
  \end{tabular*}
\end{table*}

The phase-augmented state, control, dynamics, and cost fit directly into the MPPI formulation of Eq.~\eqref{eq:mppi}.
The plan spans a horizon of $H$ control steps, but the optimization variables are $H_{\mathrm{node}}$ control nodes, from which the control signal $\tilde{\mathbf{u}}_{k:k+H}$ over the horizon is generated by linear interpolation.
This reduces the dimension of the sampled noise.
We use the joint-position targets of a PD controller as the motor control $\mathbf{u}_k$, so that the plan can be initialized directly with the reference joint trajectory $\bar{\mathbf{q}}$ read at the source speed and $d\phi^{\mathrm{src}}$.
The plan is shifted by one step after each control step.

To improve the quality of the solution, we apply the update of Eq.~\eqref{eq:mppi} $N$ times per control step with the annealing scheme of DIAL-MPC~\cite{xue2024dialmpc}.
The sampling noise of iteration $i$ at horizon step $h$ is $\sigma_{i,h}=\sigma_0\,\beta^{\,i-1}\,\beta_h^{\,H-h}$, so that the variance of the rollout samples shrinks over the iterations and near-term controls are refined more than distant ones.
In addition, only the elite fraction $\kappa$ of the samples with the lowest cost is used in the update, so that outlier samples do not affect it.

\section{Results}\label{sec:results}

We evaluate our framework on two kinds of source motions and five robots.
Human motions come from a two-hour dataset~\cite{bones_seed_2026} and are retargeted to four humanoids, the Unitree G1, R1, and H1-2 and the Booster T1.
A few quadruped motion clips are retargeted to the Unitree Go1.
We retarget the two-hour dataset with DTMR to build a temporally optimized and dynamically feasible motion dataset, and show that it serves downstream imitation learning better than the datasets produced by other motion retargeting methods.
We then show that allowing more temporal deformation yields more precise retargeting, and that DTMR achieves higher precision than the baseline under the same deformation budget.
Finally, we deploy policies trained on our references on a real robot.

\begin{table*}[t]
  \centering
  \caption{Retargeting error of DTMR compared with the temporal retargeting baseline (STMR)~\cite{yoon2025spatio}. We measure the DTW retargeting error on ten G1 clips and two Go1 clips. The error is given in mm, and the error rate in \% of the reference root path length as mean (sd) over five seeds.}
  \label{tab:prev_stmr_judge}
  \footnotesize
  \setlength{\tabcolsep}{1pt}
  \begin{tabular*}{\textwidth}{@{\extracolsep{\fill}}l c c c c c c c c c c c c@{}}
    \toprule
    & \multicolumn{5}{c}{\textit{G1, dynamic}} & \multicolumn{5}{c}{\textit{G1, static}} & \multicolumn{2}{c}{\textit{Go1}} \\
    \cmidrule(lr){2-6}\cmidrule(lr){7-11}\cmidrule(lr){12-13}
    & Jump-1 & Jump-2 & Jump-3 & Flip & Dodge & Checking & Praying & Brushing & Catch-1 & Catch-2 & HopTurn & SideSteps \\
    \midrule
    \multicolumn{13}{@{}l}{\textbf{Error (mm)} $\downarrow$} \\
    STMR~\cite{yoon2025spatio} & 95.8\,(30.4) & 85.5\,(20.6) & 556.0\,(241.6) & 1213.6\,(157.8) & 89.3\,(10.5) & 17.7\,(1.5) & 13.5\,(2.3) & \textbf{11.6}\,(1.2) & 165.3\,(130.5) & 22.8\,(0.7) & 62.7\,(8.0) & 83.2\,(13.4) \\
    DTMR (\ours) & \textbf{66.5}\,(26.5) & \textbf{40.7}\,(7.6) & \textbf{82.2}\,(17.1) & \textbf{111.6}\,(35.2) & \textbf{75.0}\,(17.1) & \textbf{14.3}\,(2.4) & \textbf{12.0}\,(1.9) & 12.3\,(1.5) & \textbf{55.3}\,(1.2) & \textbf{20.5}\,(0.9) & \textbf{57.9}\,(9.3) & \textbf{76.7}\,(9.0) \\
    \midrule
    \multicolumn{13}{@{}l}{\textbf{Error rate (\%)} $\downarrow$} \\
    STMR~\cite{yoon2025spatio} & 6.2\,(2.0) & 7.3\,(1.8) & 39.1\,(17.0) & 16.4\,(2.1) & 4.2\,(0.5) & 68.5\,(5.6) & 34.9\,(6.1) & \textbf{21.0}\,(2.2) & 4.4\,(3.5) & 3.4\,(0.1) & 4.7\,(0.6) & 4.4\,(0.7) \\
    DTMR (\ours) & \textbf{4.3}\,(1.7) & \textbf{3.5}\,(0.7) & \textbf{5.8}\,(1.2) & \textbf{1.5}\,(0.5) & \textbf{3.5}\,(0.8) & \textbf{55.3}\,(9.2) & \textbf{31.0}\,(4.8) & 22.2\,(2.8) & \textbf{1.5}\,(0.0) & \textbf{3.1}\,(0.1) & \textbf{4.4}\,(0.7) & \textbf{4.0}\,(0.5) \\
    \bottomrule
  \end{tabular*}
\end{table*}

\subsection{Dataset Construction with DTMR}
\label{subsec:dataset}
As summarized in Fig.~\ref{fig:seed2h_classes}, we retarget 999 clips of human motion, 120 minutes in total, to each of the four humanoids at the dynamics level.
This scale is possible because DTMR searches the timing and the control together in a single optimization, whose rollouts run in parallel on a GPU.
The clips are randomly sampled from BONES-SEED~\cite{bones_seed_2026}, and Fig.~\ref{fig:seed2h_donut} illustrates their motion classes, assigned from the clip names.
Locomotion (walking, jogging and running, jumping, crawling and crouching) accounts for 496 of the 999 clips, and the rest are dance, gesture and object interaction, and idle or seated motions.

Each clip is first retargeted at the kinematic level with the SoMa retargeter (SoMa-RT)~\cite{saito2026soma} and then dynamically retargeted with DTMR.
We implement the solver on MuJoCo Warp~\cite{todorov2012mujoco,mujoco_warp}, rolling out the $N_W$ samples in parallel on one GPU.
Table~\ref{tab:hparams} lists the settings, where the joint targets and the sampling noise are given in normalized joint-range units.
Refining one clip costs roughly $80$--$100\times$ its duration on an RTX~4090, about 200 GPU-hours for the two-hour corpus used below.

As illustrated in Fig.~\ref{fig:seed2h_retiming}, DTMR re-times each motion class differently to fit the dynamics of the robot.
The more dynamic a class is, the more it is slowed down, while static classes are sped up slightly.
Jogging and running are stretched on every robot (1.13--1.42$\times$ at the median), idle and seated clips are compressed (0.90--0.95$\times$), and jumps keep their length (0.96--1.03$\times$) because the slowed take-off is repaid by a hurried recovery.
In total, the corpus changes in length by at most 9.5\% (H1-2) and stays within 1\% of the source on the G1, so DTMR mostly redistributes time within the clips rather than lengthening them.

\subsection{Evaluating Improvements over Baseline Motion Retargeters}
\label{subsec:eval_baselines}

We evaluate the dynamic feasibility of the motions generated by DTMR compared with those of motion retargeting baselines.
Following prior work (e.g.,~\cite{yoon2025spatio}), we train a policy to track the retargeted motion and take its tracking error as the retargeting error.
The underlying intuition is that a motion aligned with the dynamics of the robot is readily trackable by the robot.

We use the full two-hour dataset of Sec.~\ref{subsec:dataset} and the four humanoids (G1, R1, H1-2, and T1).
The baselines are GMR~\cite{araujo2025gmr}, PHUMA~\cite{lee2025phuma}, OmniRetarget~\cite{yang2025omniretarget}, and SoMa-RT~\cite{saito2026soma}.
Since the dataset contains roughly a thousand clips, we train a single off-the-shelf general motion tracking policy~\cite{ze2025twist} on the whole dataset instead of a separate policy for each clip.
One policy is trained per robot and per retargeter with identical hyperparameters.
We measure the retargeting error by the global mean per-joint position error (MPJPE-G) of the policy over the training clips, which represents the dynamic feasibility of the retargeted motion.
We also measure the error after dynamic time warping (DTW), which is insensitive to temporal deformation and thus represents how well the semantics of the source motion are preserved.
We call it the DTW retargeting error.
Although our method can be applied on top of any of these baselines, we apply it on top of SoMa-RT.
We report the $25\%$, $50\%$, and $75\%$ quartiles of the per-clip retargeting error, denoted $Q_{25}$, $Q_{50}$, and $Q_{75}$.
This separates the clips by tracking difficulty, as $Q_{25}$ reflects the static motions that are simple to track and $Q_{75}$ the dynamic motions that are difficult to track.

Table~\ref{tab:twist_training} reports the dynamic feasibility of the retargeted motions.
It shows that dense temporal optimization improves the static motions modestly and the dynamic motions substantially.
Our method achieves the lowest error in every column except $Q_{25}$ on the R1 and the H1-2, where SoMa-RT is marginally ahead.
In particular, compared with SoMa-RT alone, our method reduces the median error by 13.4\% on average across the four robots (21.5\%, 7.5\%, 10.2\%, and 14.5\% on the G1, R1, H1-2, and T1).
The improvement is larger on dynamic motions.
Compared with SoMa-RT, our method improves $Q_{25}$ by only 1.8\% on average but $Q_{75}$ by 14.9\%.
Against the average of all four baselines, the improvement is 15.1\% at $Q_{25}$ and 43.4\% at $Q_{75}$.

Table~\ref{tab:twist_dtw} reports the semantic fidelity of the retargeted motions by the DTW retargeting error.
The overall trend is similar to Table~\ref{tab:twist_training}.
Our method has the lowest error in every column except $Q_{25}$ on the H1-2, and it reduces the median error of SoMa-RT by 12.9\% on average across the four robots (20.5\%, 7.2\%, 9.8\%, and 14.0\% on the G1, R1, H1-2, and T1).
The gain again concentrates on the dynamic clips: 2.2\% at $Q_{25}$ but 11.4\% at $Q_{75}$ against SoMa-RT, and 15.8\% at $Q_{25}$ but 39.8\% at $Q_{75}$ against the average of the four baselines.
Temporal optimization thus benefits both, modestly for static motions and substantially for dynamic motions.

\subsection{Evaluating Improvements over the Temporal Retargeting Baseline}
\label{subsec:prev_stmr}

We aim to compare the proposed DTMR against a baseline that also optimizes the temporal dimension of the motion.
In particular, we compare against STMR~\cite{yoon2025spatio}.
Both STMR and DTMR start from the same kinematic retargeting with SoMa-RT~\cite{saito2026soma}.
However, there are two main differences between the two methods.
First, STMR optimizes the temporal dimension in coarse motion segments, while DTMR optimizes it densely at every frame.
Second, STMR performs a nested optimization, while DTMR optimizes the temporal parameters and the control in a single optimization.
Specifically, STMR searches the temporal parameters with Bayesian optimization in the outer loop and finds the control with an inner controller.
Notably, STMR uses iLQG~\cite{tassa2012synthesis} as the inner controller.
We replace it with our MPPI tracker without the phase variable so that the rollouts run in parallel on a GPU.

We adopt the DTW retargeting error as in Sec.~\ref{subsec:eval_baselines} and experiment with both a humanoid and a quadruped robot.
We use only a few motion clips, since the nested optimization of STMR is slow.
In more detail, we use ten clips from the dataset of Sec.~\ref{subsec:dataset} with the Unitree G1, five dynamic and five static, and the two quadruped motions (HopTurn and SideSteps) used in~\cite{yoon2025spatio} with the Unitree Go1.
For each retargeted motion, a DeepMimic~\cite{peng2018deepmimic} policy is trained (five seeds) and its DTW retargeting error is reported.

\begin{figure}[!tb]
  \centering
  \begin{subfigure}[b]{\linewidth}
    \includegraphics[width=\linewidth]{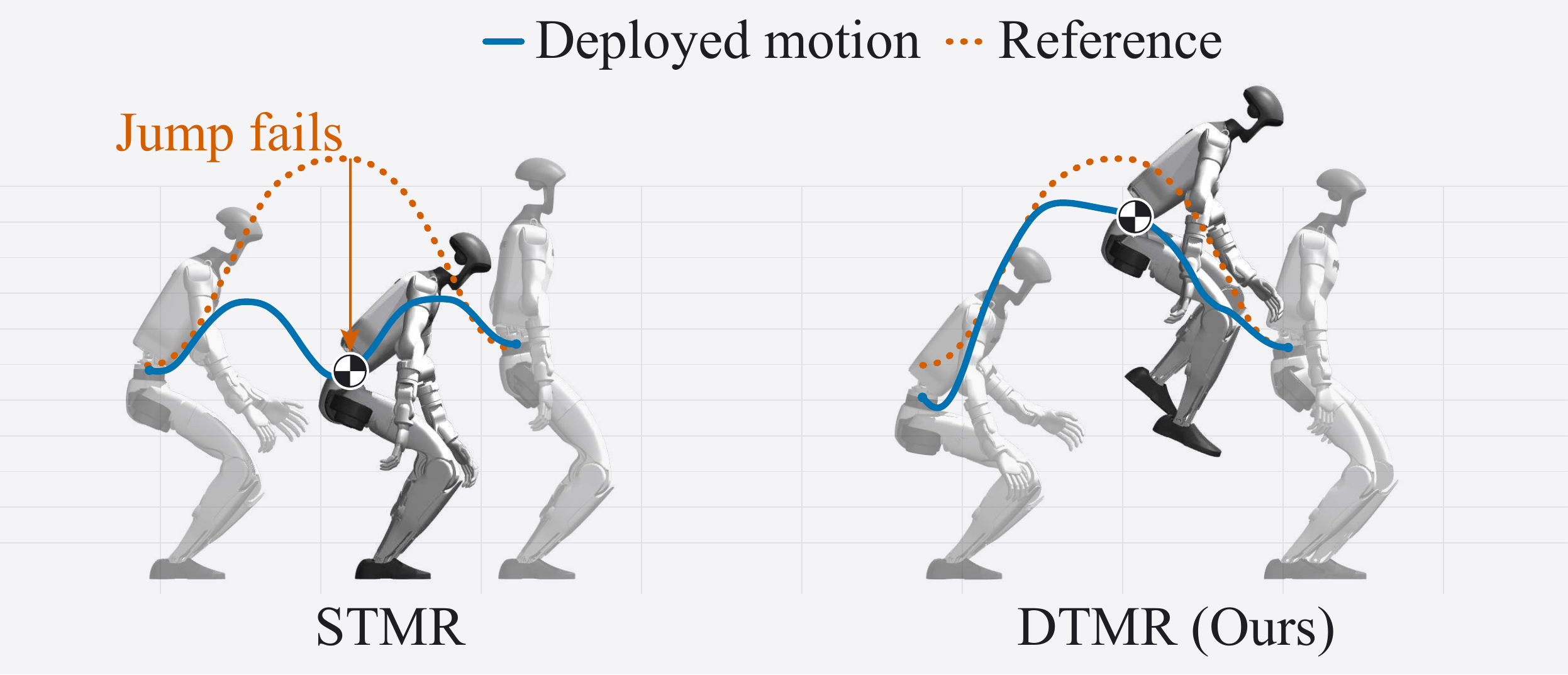}
    \caption{High Jump}
    \label{fig:qual_vs_stmr_jump}
  \end{subfigure}
  \\[0.6em]
  \begin{subfigure}[b]{\linewidth}
    \includegraphics[width=\linewidth]{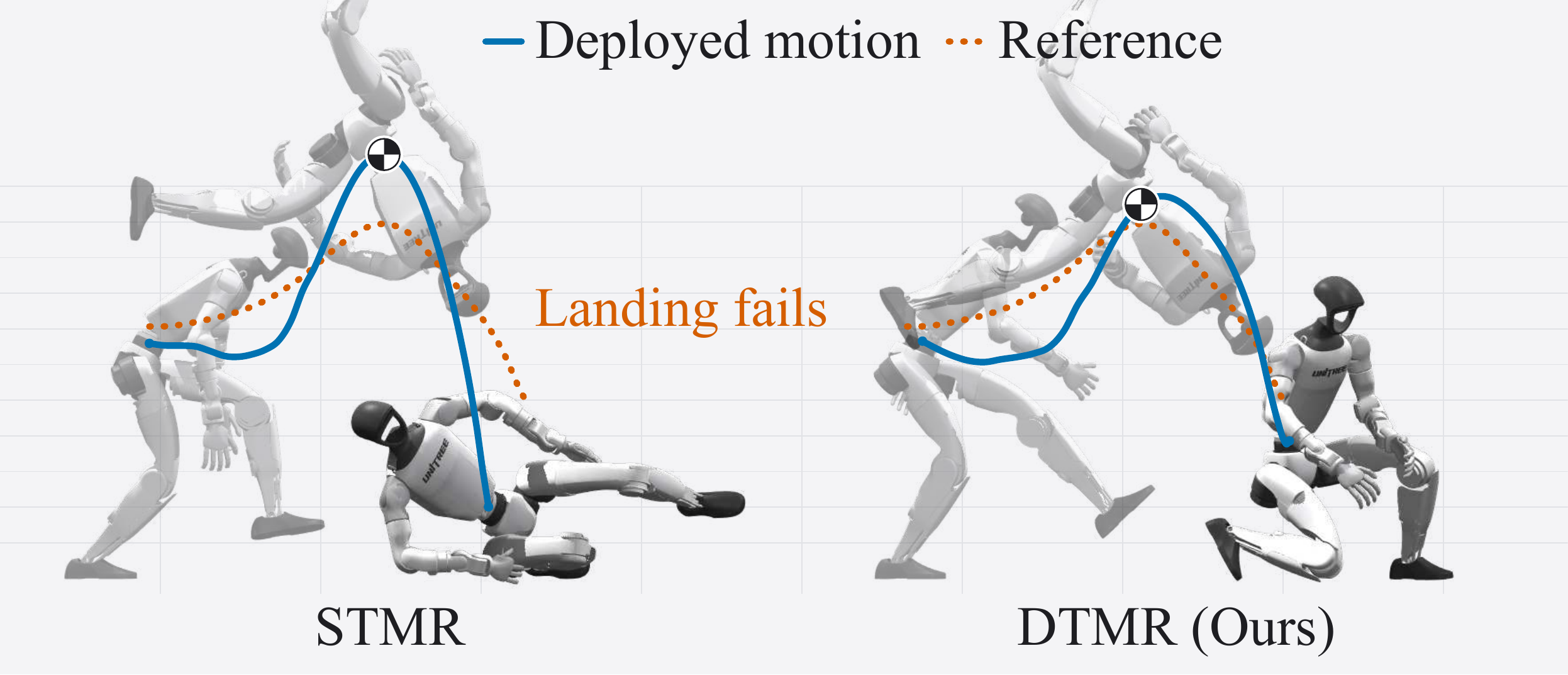}
    \caption{Flip}
    \label{fig:qual_vs_stmr_flip}
  \end{subfigure}
  \caption{DTMR retargets motions with flight phases more successfully than the temporal retargeting baseline (STMR)~\cite{yoon2025spatio}.
    In these clips, (a) STMR skips the jump and (b) fails the landing, whereas DTMR completes both.}
  \label{fig:qual_vs_stmr}
\end{figure}

\begin{table}[!tb]
  \centering
  \caption{The wall-clock time to retarget the clips is compared between DTMR and STMR~\cite{yoon2025spatio} on the same GPU and tracker.}
  \label{tab:prev_stmr_compute}
  \small
  \setlength{\tabcolsep}{4pt}
  \begin{tabular*}{\linewidth}{@{\extracolsep{\fill}}l rrr@{}}
    \toprule
    Motion set            & STMR~\cite{yoon2025spatio} & DTMR (\ours) & speedup \\
    \midrule
    G1 (10 clips, total)  & 45.4\,h      & \textbf{2.4\,h}  & $19.2\times$          \\
    \quad per clip, mean  & 4.5\,h       & \textbf{14\,min} & $\mathbf{18.9\times}$ \\
    \quad per clip, range & 2.0--11.8\,h & \textbf{7--34\,min} & $\mathbf{17.1\text{--}20.7\times}$ \\
    \midrule
    Go1 (2 clips, total)  & 2.4\,h       & \textbf{7\,min}  & $20.8\times$          \\
    \quad per clip, mean  & 73\,min      & \textbf{3.5\,min} & $\mathbf{20.5\times}$ \\
    \quad per clip, range & 53--94\,min  & \textbf{3--4\,min} & $\mathbf{18.9\text{--}22.2\times}$ \\
    \bottomrule
  \end{tabular*}
\end{table}

\begin{figure}[t]
  \centering
  \includegraphics[width=\columnwidth]{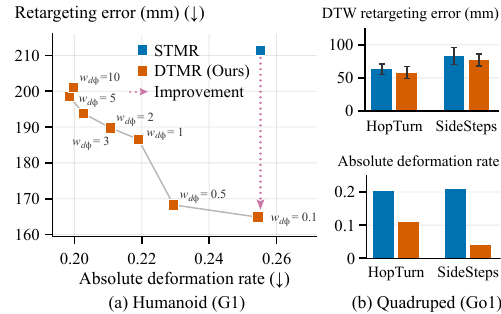}
  \caption{DTMR retargets precisely by deforming only the parts of the motion that need a change, on (a) a humanoid and (b) a quadruped. (a) On the G1, the retargeting error decreases as DTMR allows more temporal deformation, and DTMR is more precise than the baseline (STMR)~\cite{yoon2025spatio} at the same absolute deformation rate $\langle|r|\rangle$. (b) On the Go1, DTMR reaches a similar error with a smaller absolute deformation rate.}
  \label{fig:dphi_tradeoff}
\end{figure}

As shown in Table~\ref{tab:prev_stmr_judge}, DTMR achieves a lower DTW error than STMR on nine of the ten G1 clips.
The dynamic clips move the pelvis fast (up to 3.4\,m/s $\bf v_{xy}$ and 2.0\,m/s $\bf v_{z}$), whereas the static clips stay below 0.6 and 0.3\,m/s.
The error is reduced by 55.1\% on average on the dynamic clips and by 20\% on the static clips, which shows that dense temporal optimization is better than optimizing in coarse segments where the timing matters.
On three clips the slowed STMR reference breaks the tracked policy: it falls at the landing of Flip, ignores the jump in Turn-Jump $135^\circ$, and falls in half of the seeds of Big-Fish Catch.
Fig.~\ref{fig:qual_vs_stmr} shows a qualitative example on a G1 jump, where STMR lowers the jump height whereas DTMR preserves it.
On the Go1 (Fig.~\ref{fig:dphi_tradeoff}(b)), DTMR reaches a slightly lower error than STMR on both clips.
Moreover, as Table~\ref{tab:prev_stmr_compute} shows, DTMR is ${\sim}19\times$ faster than STMR on the G1 and ${\sim}21\times$ on the Go1, since it solves the timing and the control in a single optimization rather than in a nested one.

\begin{figure}[!t]
  \centering  \includegraphics[width=\linewidth]{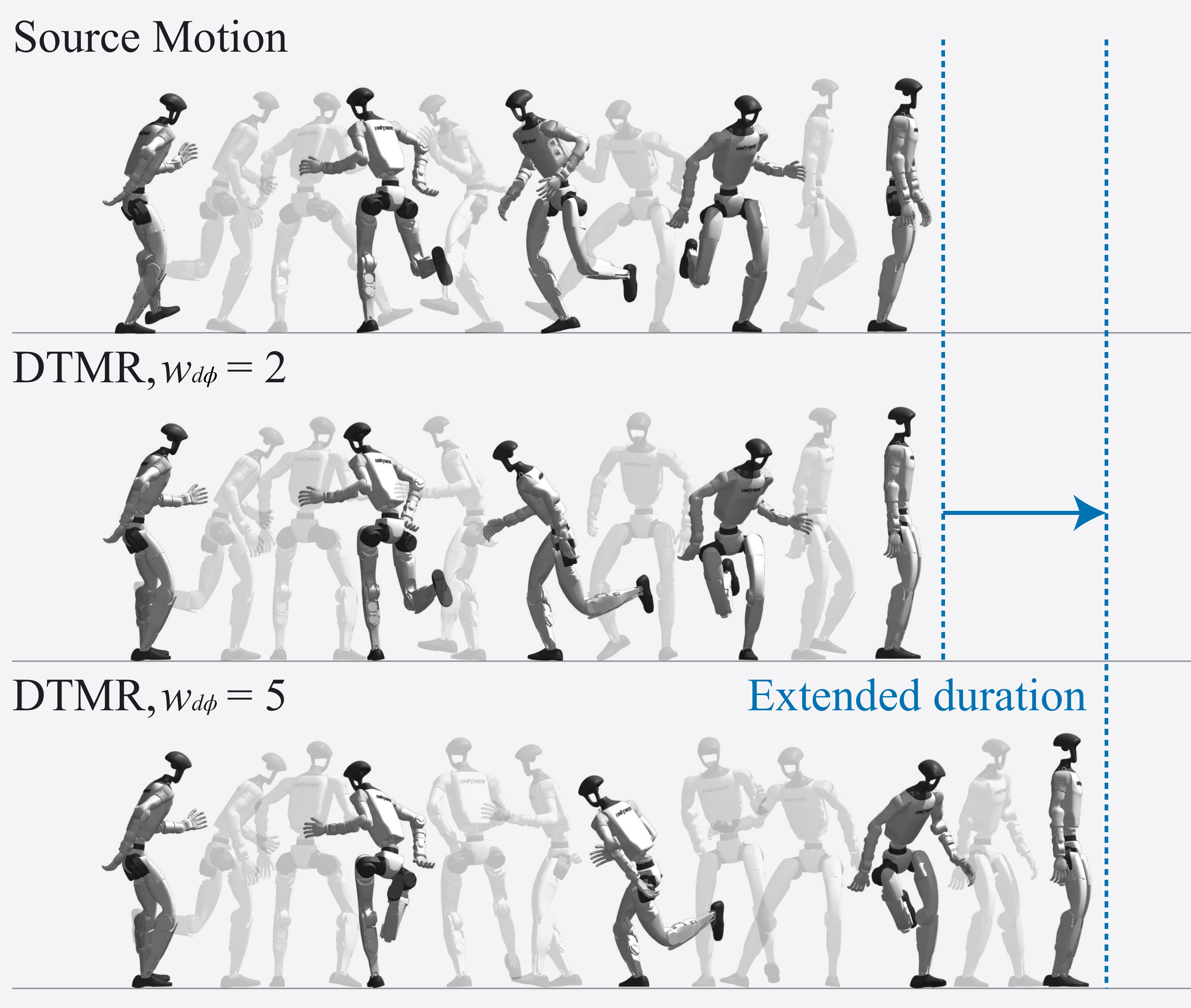}
  \caption{DTMR can control how much the timing of a motion is deformed. A large $w_{d\phi}$ keeps the timing close to the source motion, and a small $w_{d\phi}$ deforms it more so that the robot follows the motion more precisely.}
  \label{fig:dance_wdphi}
\end{figure}

\subsection{Evaluating the Deformation--Precision Trade-off}
\label{subsec:eval_tradeoff}

We evaluate the controllability of our method over the trade-off between temporal deformation and precision.
In detail, the amount of temporal deformation is controlled by the phase-cost weight $w_{d\phi}$ of Eq.~\eqref{eq:cost}.
A large $w_{d\phi}$ keeps the source timing, and a small one allows more deformation.
Fig.~\ref{fig:dance_wdphi} illustrates this on a dance clip, where the motion is slowed down as $w_{d\phi}$ decreases.
We sweep $w_{d\phi}$ on ten clips labeled as jumps in the dataset and score the retargeted motions with a downstream tracking policy.
We measure the amount of temporal deformation by the absolute deformation rate $\langle|r|\rangle=\tfrac{1}{T_\phi}\sum_k|r_k|$, the mean of $|r_k|$ over the rollout.

\begin{figure}[!tb]
  \centering
  \begin{subfigure}[b]{0.49\linewidth}
    \includegraphics[width=0.99\linewidth]{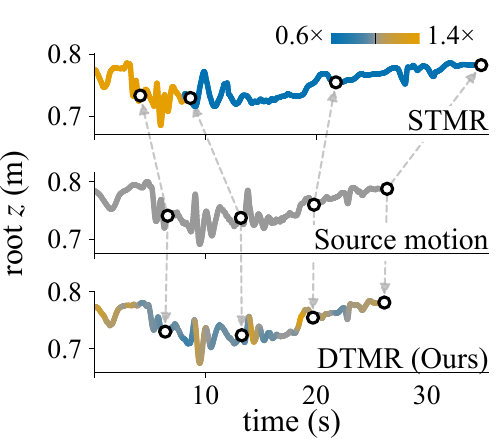}
    \caption{Big-Fish Catch (static).}
    \label{fig:bigfish_timeline}
  \end{subfigure}
  \begin{subfigure}[b]{0.49\linewidth}
    \includegraphics[width=0.99\linewidth]{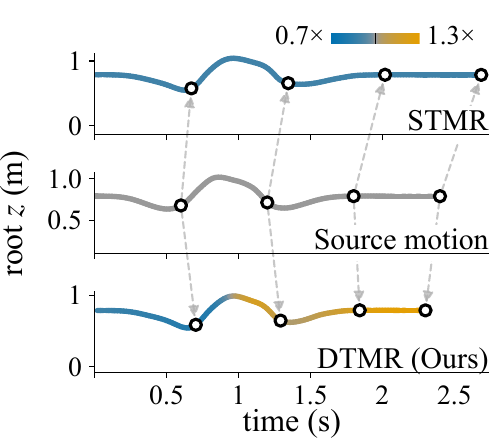}
    \caption{High-Jump (dynamic).}
    \label{fig:highjumpA340_timeline}
  \end{subfigure}
  \caption{DTMR optimizes the timing densely, while the baseline~\cite{yoon2025spatio} scales coarse segments. The source motion (middle) is retimed by the baseline (top) and by DTMR (bottom), and the colour of the curve is the deformation rate.}
  \label{fig:g1_timelines}
\end{figure}

\begin{figure}[!t]
  \centering
  \begin{subfigure}[b]{\linewidth}
    \centering
    \includegraphics[width=0.9\linewidth]{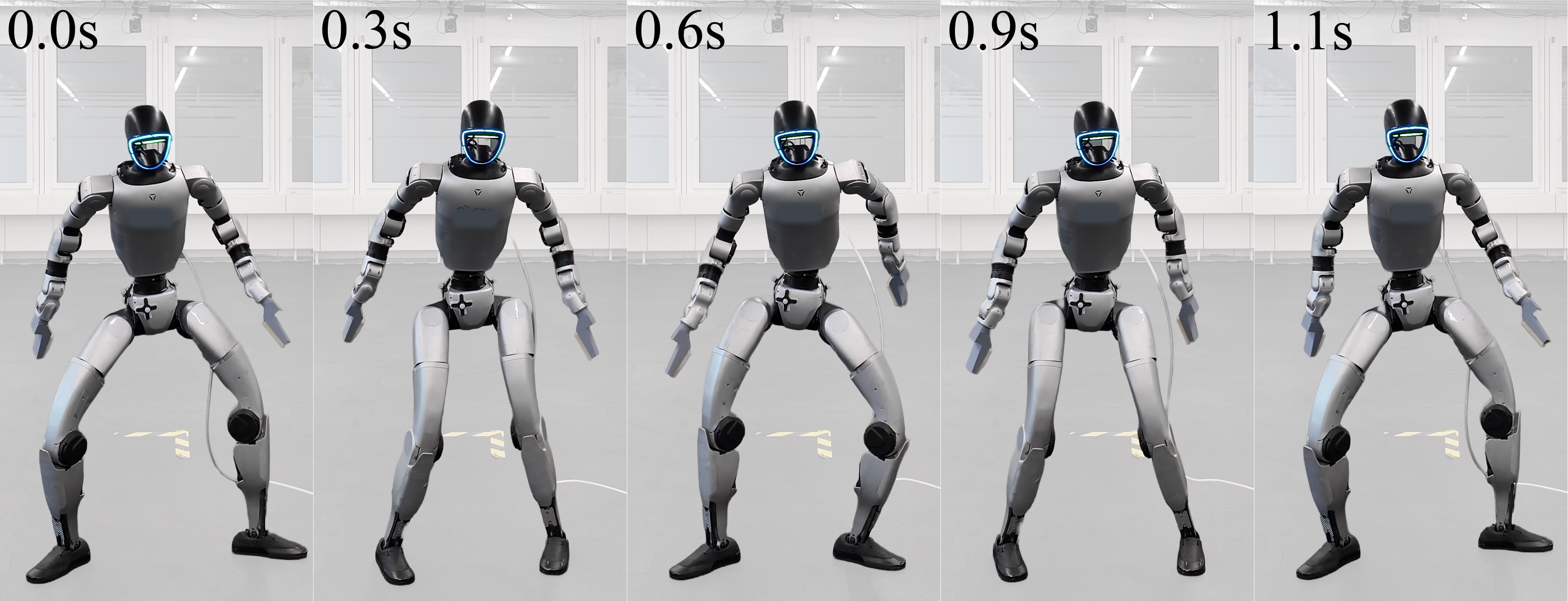}
    \caption{ Foot-shuffle dance}
    \label{fig:realworld_frames}
  \end{subfigure}
  \\[0.6em]
  \begin{subfigure}[b]{\linewidth}
    \centering
    \includegraphics[width=0.9\linewidth]{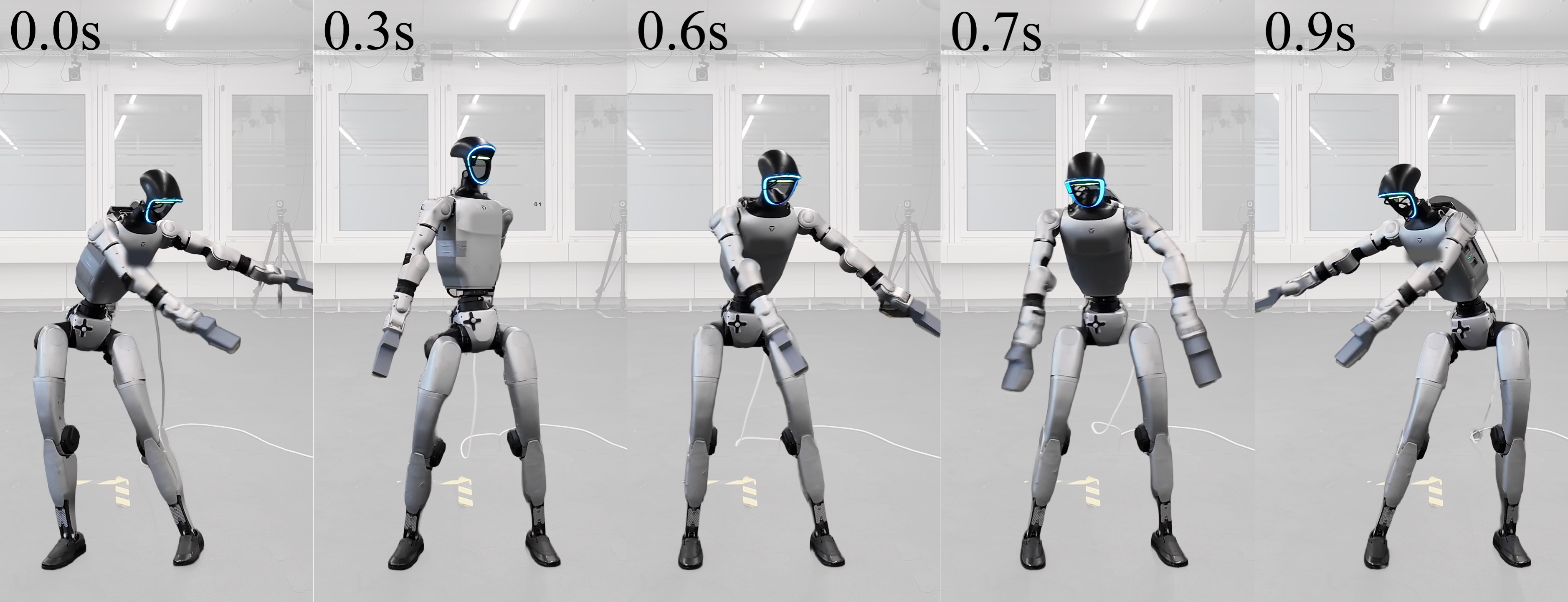}
    \caption{ Weight-shift dance}
    \label{fig:realworld_frames2}
  \end{subfigure}
  \caption{Policies trained on DTMR references are deployed on a Unitree G1 on (a) a contact-rich dance and (b) a dance with fast center-of-mass motion.}
  \label{fig:realworld}
\end{figure}

As shown in Fig.~\ref{fig:dphi_tradeoff}, allowing more temporal deformation improves precision.
The error falls monotonically from 202\,mm at $w_{d\phi}=10$ to 170\,mm at $w_{d\phi}=0.5$, and the settings with $w_{d\phi}\le0.5$ all score 164--170\,mm.
Reducing $w_{d\phi}$ from 5 to 0.5 increases $\langle|r|\rangle$ from 0.20 to 0.23 and lowers the retargeting error by 30\,mm.

Moreover, DTMR is more precise than STMR under the same temporal deformation budget.
We run STMR on the same clips and measure its absolute deformation rate $\langle|r|\rangle$.
In Fig.~\ref{fig:dphi_tradeoff}, STMR is marked as a single point at its measured deformation.
As shown in Fig.~\ref{fig:dphi_tradeoff}, DTMR is 22\% more precise than STMR at the same deformation ($w_{d\phi}=0.1$), and still 5\% more precise with 22\% less deformation ($w_{d\phi}=5$).
The Go1 clips of Table~\ref{tab:prev_stmr_judge} show the same trend: DTMR has an absolute deformation rate of $0.11$ on HopTurn and $0.04$ on SideSteps, against 0.20 and 0.21 for STMR, so it matches the precision of STMR with 2--5$\times$ less deformation.

\subsection{Evaluating the Timing Adjustment}
\label{subsec:eval_temporal}

We evaluate how densely our method deforms the timing of each motion.
Fig.~\ref{fig:teaser} shows a high jump retargeted to the G1.
DTMR delays the onset of the jump and shortens the flight phase so that the robot can follow the motion, while the rest of the motion keeps the source timing.
Such localized changes are possible because DTMR optimizes the deformation rate at every step of the motion rather than once per segment.

We further compare the deformation with that of STMR, which optimizes the timing in coarse motion segments.
We retarget a static clip and a jump of the G1 and record the height of the pelvis, as shown in Fig.~\ref{fig:g1_timelines}.
Compared with the kinematic reference, STMR applies one deformation rate to each segment of the motion and lengthens it by 32\% and 12\% on the static and the dynamic clip, since a slower motion is easier to track.
In contrast, DTMR keeps the static motion at the source tempo and deforms the timing only around the jump, where it slows down the crouch and speeds up the landing.

\subsection{Real-World Deployment}
\label{subsec:eval_real}
We deploy policies trained on DTMR references with BeyondMimic~\cite{liao2025beyondmimic} on a Unitree G1, as shown in Fig.~\ref{fig:realworld}.
Figs.~\ref{fig:realworld}(a) and \ref{fig:realworld}(b) represent a contact-rich dance that shuffles the feet on the ground and a dance that moves the center of mass quickly, respectively.
DTMR re-times these dances according to the dynamic properties of the G1, which makes them straightforward for the robot to imitate.
On the weight-shift dance, the arm sweeps are slowed to $0.84\times$ the source speed.
Training on our references also yields a higher final tracking reward than training on the kinematic retargeting alone.
The tracking reward sums the per-step rewards on the tracking error of the joints, the links, and the root, and increases as the policy follows the reference more closely.
On the two deployed dances, the final mean tracking reward rises from 40.8 to 42.5 on the foot-shuffle dance and from 38.9 to 41.7 on the weight-shift dance.
This indicates that dynamically feasible references are easier to learn.

\section{Conclusion}\label{sec:conclusion}

We presented DTMR, which jointly optimizes the timing and the control of every frame within a single optimization.
As a result, only the parts of the motion that need a change in timing are deformed, and the rest keeps the source timing.
We showed that dense timing adjustment helps dynamic motions the most and that the amount of temporal deformation gives a knob between timing preservation and precision.
Under the same deformation budget, DTMR retargets more precisely than the baseline and ${\sim}19\times$ faster.
We retargeted a two-hour human motion dataset to four humanoids, and policies trained on it learned dynamic motions better and transferred to a real robot.

Nevertheless, several limitations exist.
First, DTMR fits the timing to the dynamic properties of the robot, not to the meaning of the motion.
A dance loses its meaning when it falls out of time with the music.
Our framework partially mitigates this by exposing the phase-cost weight $w_{d\phi}$, which bounds how much the timing may change (Sec.~\ref{subsec:eval_tradeoff}), but it does not know which parts of the motion must keep their timing.
Second, DTMR does not handle object interaction, where the timing of a contact must be adjusted together with the motion of the object.
As future work, we plan to include the state of the object in the optimization so that the retargeted timing of the robot and the motion of the object are decided together.

\bibliographystyle{IEEEtran}
\bibliography{citations}

\end{document}